\documentclass{sigkddExp}

\usepackage{graphicx}
\usepackage{amsmath}
\usepackage[caption=false]{subfig}
\graphicspath{ {images/} }
\usepackage[labelfont=bf,textfont=bf]{caption}
\usepackage{makecell}
\usepackage{flushend}
\usepackage{cite}
\numberofauthors{4}
\author{Loveperteek Singh\\Myntra Designs, India\\love.perteek@myntra.com
\\[3ex] Sagar Arora\\Myntra Designs, India\\sagar.arora@myntra.com 
\and Shreya Singh\\Myntra Designs, India\\shreya.singh1@myntra.com 
\\[3ex] Sumit Borar\\Google, India\\sumitborar@gmail.com\thanks{Work done while at Myntra}}

\title{One Embedding To Do Them All}

\begin{document}
\maketitle

\begin{abstract}

Online shopping caters to the needs of millions of users daily. Search, recommendations, personalization have become essential building blocks for serving customer needs. Efficacy of such systems is dependent on a thorough understanding of products and their representation. Multiple information sources and data types provide a complete picture of the product on the platform. While each of these tasks shares some common characteristics, typically product embeddings are trained and used in isolation.

In this paper, we propose a framework to combine multiple data sources and learn unified embeddings for products on our e-commerce platform. Our product embeddings are built from three types of data sources - catalog text data, a user's clickstream session data and product images. We use various techniques like denoising auto-encoders for text, Bayesian personalized ranking (BPR) for clickstream data, Siamese neural network architecture for image data and combined ensemble over the above methods for unified embeddings. Further, we compare and analyze the performance of these embeddings across three unrelated real-world e-commerce tasks specifically checking product attribute coverage, finding similar products and predicting returns. We show that unified product embeddings perform uniformly well across all these tasks.




\end{abstract}


\section{Introduction}


E-commerce is growing at a phenomenal rate around the world. Matching consumer's need and retrieving relevant products is pivotal to the business. This has led to a lot of research in areas of search, recommendation systems, personalization, demand prediction etc. For all these tasks, detailed understanding of product and users become extremely important. Users are typically represented by their activity on the portal like clicks, product viewed, purchases etc and the explicit information provided by them. Explicit information is typically noisy and less reliable. On the other hand, products are usually well defined with cleaner titles, attributes, cataloged images, description, reviews etc. Products representation is crucial in solving for all these tasks. Typically, each task is solved in isolation. Furthermore, each task involves understanding the product via \textit{embeddings}. 

However, all products share some common semantic information; irrespective of the downstream task involved. This makes it possible to create a common product knowledge layer in the form of \textit{unified product embeddings}. For instance, a premium t-shirt from BOSS brand might be recommended to niche segment of people who have an intent in similar products, can demonstrate low return rates and comparatively lower demand too. If an embedding could capture such semantics like brand information and the notion of premiumness of a product, it can be utilized across varied e-commerce tasks.

In this paper, we use the following data sources to represent a product:

\begin{enumerate}
    \item \textbf{Textual Data}: This involves products' title (name), description and cataloged attributes like brand, color, fabric and physical attributes like neck, pattern etc. Product titles are structured and the average length of product title is 7.3 words. Product descriptions vary a lot based on the products and contain both structured and unstructured information. Hereafter we used the phrases \textit{textual data} and \textit{side information} interchangeably.
    \item \textbf{Clickstream Data}: This includes all the users' sessions and the involved interactions including searches, impressions, clicks, sorts and, filters used, add to carts, purchases etc. These signals are good indicators for visibility and popularity of products on the platform.  
    \item \textbf{Visual Data}: This includes product images available in the catalog. Each product on an average is represented by at least 4 images. These images are mostly shot in a controlled setting with solid color background and model poses. 
\end{enumerate}

Our work focuses on capturing a wider variety of signals from various data sources (as mentioned above) to embed all products in a \textit{product semantics} space that can potentially tackle a wide variety of e-commerce problems. We create product embeddings from different data sources in isolation (for example we use Bayesian Personalised Ranking to embed products using only clickstream data, denoising autoencoder for textual data and siamese network for the visual data) and finally use an ensemble over skip-gram based architecture and siamese architecture to collate multiple data sources to create \textit{unified product embedding}. We compare and contrast our unified embeddings on three different tasks:
\begin{enumerate}
\item \textbf{Embedding to Attribute}: This task attempts to evaluate learned embeddings on how well they can capture the products' textual attributes like brand, color etc.
    \item \textbf{Clicked-Purchased Product Similarity}: we compute the similarity of the purchased product in a session with those which were clicked. We show how our unified embeddings are able to better capture the similarity.
    \item \textbf{Cart Return Prediction}: Returns ensue bad user experience apart from extra operational costs incurred by our platform.  Hence, through cart return prediction,  we aim to identify the cart products which have a high probability of being returned and take corrective actions. This task involves using product embeddings to predict if a user $u$ would return a product $p$.
    
\end{enumerate}

The rest of the paper is organized as follows. Section \ref{related} discusses the related work and previous attempts to collate multiple data sources. Section \ref{methodology} discusses various approaches to creating product embeddings using different combinations of data sources and our approach to create unified embeddings. Finally, we evaluate our embeddings on two different downstream tasks in Section \ref{results}.

\section{Related Work} 
\label{related}
Traditionally, product representations have been learned through Matrix Factorization and related approaches\cite{rendle2009bpr, hu2008collaborative} which use only user's feedback. For implicit feedback setting, interpreting unobserved feedback poses a challenge. \cite{hu2008collaborative} interprets unobserved feedback to be negative thereby associating weights with feedback and factorize the resultant weighted matrix. \cite{rendle2009bpr} proposed BPR-MF which optimizes directly for the ranking of the feedback. These embeddings are less useful for cold-start settings, long-tailed datasets and are very specific to capture only implicit feedback signals thus limiting their utility to personalization/recommendation tasks only.

In Fashion domain, \cite{he2016ups, he2016vbpr, kang2017visually} have successfully tried to incorporate image embedding from deep CNN's of various architectures. Along with objectives like modeling fashion trends\cite{he2016ups} and generating images using GAN's\cite{kang2017visually}, all of them are shown to perform well on personalized recommendations task only.
\cite{he2016ups} learn product embeddings using image and clickstream data but do not use side information like catalog data. Also, they evaluate their embeddings on personalization task only whereas we focus on wide variety of tasks relevant to e-commerce business.

More recently Neural Network based approaches inspired from Word2Vec \cite{mikolov2013efficient} have been proposed for learning product representations for Recommender systems. The intuition behind Word2Vec, the distributional hypothesis\cite{sahlgren2008distributional} is \textit{words that occur in the same contexts tend to have similar meanings}. Similar hypothesis \textit{users inclined towards a similar set of products have similar underlying tastes} is the basis of collaborative filtering. \cite{levy2014neural} shows the similarity between matrix factorization and Word2Vec based approaches.

Word embedding based approaches include Prod2Vec \cite{grbovic2015commerce} and
MetaProd2Vec \cite{vasile2016meta}. \cite{grbovic2018real} have learnt product embeddings using clickstream and transactional data. 
Prod2vec uses co-occurrence of products to learn product embeddings whereas MetaPord2Vec also uses side information for learning. While there has been work on incorporating product description, side information and images into Recommender systems in isolation, only \cite{nedelec2017specializing} has proposed to incorporate description and images along with clickstream data into Recommender system model which has been evaluated on recommendation tasks.


Another novel way to learn embeddings is from graph-based learning techniques such as DeepWalk\cite{perozzi2014deepwalk}, LINE\cite{tang2015line} and Node2Vec\cite{grover2016node2vec}. We also compare performance embeddings learned from DeepWalk on our item co-purchase, co-browse graph.

While most of the work has focused on using all or a subset of data sources for recommendation system settings, our work focuses on building a unified (using all possible data sources) product representation which can form a fundamental layer for training and fine-tuning models for different e-commerce tasks ranging from recommendations to return prediction.

\section{Methodology} \label{methodology}

\begin{figure}[h]
    \centering
    \includegraphics[width=8cm,height=6cm]{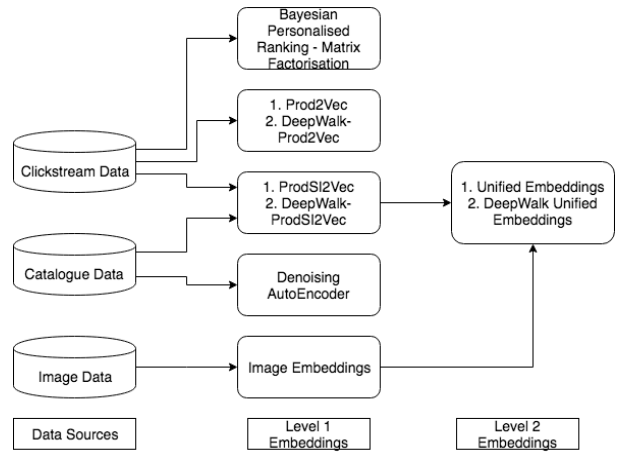}
    \caption{Different Techniques to Learn Product Embeddings}
    \label{fig:Approaches}
\end{figure}

This section describes different ways to learn product embeddings. As shown in Figure \ref{fig:Approaches} we evaluate embeddings learned from different data sources- 
\begin{enumerate}
    \item Clickstream Data: BPR-MF, Prod2Vec and DeepWalk-Prod2Vec
    \item Content Data (Catalogue and Image): Denoising Autoencoder and Image Embeddings
    \item Clickstream and Content Data: ProdSI2Vec (ProductSide\-Informa\-tion2Vec), DeepWalk-ProdSI2Vec and Unified Embeddings
\end{enumerate}

In addition to using user's lifetime data, we also compare the performance of Prod2Vec and Prod-SI2Vec with graph based embeddings learned from a platform level item-item graph. Table \ref{table:Table 1} describes the terminology used.
\begin{table}[]
\centering
\begin{tabular}{|l|l|l|l|l|l|}
\hline
\textbf{Symbol} & \textbf{Meaning}\\ \hline
\(U\) & the set of all users \\ \hline
\(P\) & the set of all products \\ \hline
\(E_c\) & click event \\ \hline
\(E_b\) & add to bag event \\ \hline
\(E_p\) & purchase event \\ \hline
\(\gamma_u\) & latent embedding of user \(u\) \\ \hline
\(\gamma_p\) & latent embedding of product \(p\) \\ \hline

\end{tabular}
\caption{Terminology Used}
\label{table:Table 1}
\end{table}

\subsection{Bayesian Personalised Ranking-MF}
Product embeddings are learned using Matrix Factorisation for modeling implicit feedback and Bayesian Personalised Ranking as a pairwise ranking optimization method as described in \cite{rendle2009bpr}. Following equation models predicting implicit feedback from user's latent embedding \(\gamma_u\)  and product's latent embedding \(\gamma_p\),
\begin{equation} 
     x_{u, p} = \alpha + \beta_u + \beta_p + \gamma_u^T\gamma_p
\end{equation}
where \(\alpha\) is global offset, \(\beta_u\) and \(\beta_p\) are bias terms.
BPR maximizes the following objective, whereby for each \(i, j\) product pair for which user \(u\) has given positive implicit feedback about product \(i\) and not observed product \(j\), 
\begin{equation} 
     \sum_{\forall {(u,i,j)}} log(\sigma(x_{u, i} - x_{u, j})) -  \lambda_\theta{||\theta||}^2
\end{equation}

where \(\theta = (\gamma_p, \gamma_u)\ \forall{(p,u) \in (P,U)}\) and \(\lambda_\theta\) are regularization parameters.

As in \cite{agarwal2018personalizing}, we construct a user-product matrix with values as total interactions (clicks, carts and, purchases) and train BPR-MF on it to learn product embeddings.

\subsection{Denoising AutoEncoder}
\begin{figure}[h]
    \centering
    \includegraphics[width=6cm,height=4cm]{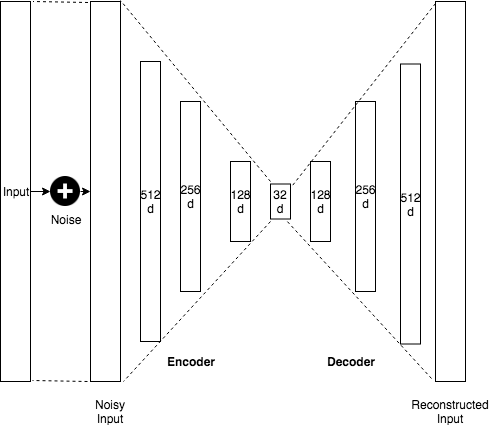}
    \caption{Autoencoder Architecture}
    \label{fig:autoenc}
\end{figure}

We train a Denoising AutoEncoder\cite{vincent2008extracting} for a pure content-based recommendation setting. All side information of products is one hot encoded and fed into a denoising autoencoder. We use a stacked denoising autoencoder architecture to encode and then decode the corrupted one hot vectors. Denoising AutoEncoder minimizes the reconstruction error between corrupted input and reconstructed input, as follows:
\begin{equation} \label{eq:imp}
     J(\theta)= \frac{1}{2|P|}\sum_{i=1}^{i=|P|}||x_i - Dec(Enc(x_{i_{corr}}))||^2
\end{equation}
where \(x_i\) is one-hot encoded vector representation of Product i, \(x_{i_{corr}}\) is input corrupted with uniform noise sampled from \([0,1]\), \(Dec()\) and \(Enc()\) are Decoder and Encoder architectures respectively.
Then the output of the encoder part is used as the latent representation of the product. As products have a large number of physical attributes, that can potentially impact how much similar two products are. We use a 3 layer deep architecture for encoder and decoder. The architecture is shown in Figure \ref{fig:autoenc}.

\subsection{Image Embeddings}
Visual aesthetics have an important role to play while understanding fashion product. Images have a significant influence on user taste. Therefore, for another content-based recommendation setting, we learn image embeddings using the architecture proposed in \cite{wang2014learning}. The deep convolution neural network architecture learned using a triplet loss has been specifically trained to learn fine-grained image similarity.   

\subsection{Word2Vec Based Approaches}
In this section, we discuss four approaches based on Word2Vec architecture. The first two approaches use co-purchased and co-bagged(added to bag) data from a user's lifetime history. The other two are based upon learning skip-gram based embeddings on simulated sessions after DeepWalk.

\subsubsection{Prod2Vec}
\label{Word2Vec:Prod2Vec}
In line with \cite{arora2016decoding}, in Prod2Vec approach, we use the lists of co-purchased and co-bagged(added to bag) products throughout the lifetime of a user on the platform sorted by timestamp. A sample list for a user \(u\) looks like:
\begin{equation}
     L_u = (b_1, b_1, ..., p_1, ..., p_2, p_3, ..., b_i, p_j)
\end{equation}
where \(b_i\) represents the \(i\)th product added to bag and \(p_j\) represents \(j\)th product purchased by the user \(u\) in his lifetime.
\begin{figure}[h]
    \centering
    \includegraphics[width=8cm,height=4cm]{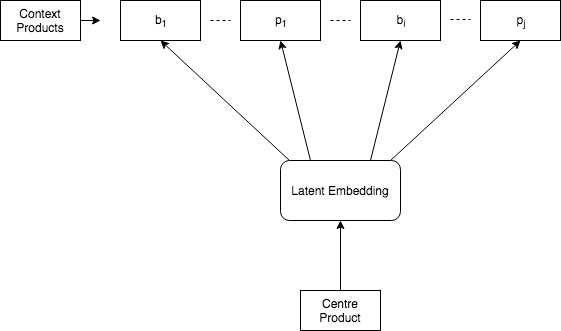}
    \caption{Prod2Vec Architecture}
    \label{fig:Prod2Vec}
\end{figure}
These lists are used as sentences in Word2Vec model. For each product(in bag and purchased) as center word in the list we sample all other product in the list as context words. This is equivalent to generating all product-product(centre-context) pairs from the list and setting window size to one.
The latent representations of the products are learned using the Skip Gram with Negative Sampling model. We sample negative samples randomly from other users' lists. In Skip Gram with Negative Sampling we maximize the log-likelihood of observing a context product given centre product as follows:
\begin{equation} \label{eq:imp}
     J(\gamma_p)= -\frac{1}{|U|}\sum_{t = 1}^{t = |U|}-\frac{1}{|L_u|}\sum_{\forall{x} \in L_u}\sum_{\forall{y} \in L_u s.t. y \neq x}\log P(y/x;\gamma_p)
\end{equation}
where x denotes the centre product, y denotes the context product and \(\gamma_p\) denotes the product embeddings.
\subsubsection{Prod-SI2Vec}
For each product we have 6 types of side information(SI) in form of key-value pairs:
\begin{enumerate}
    \item Brand:{Nike, Puma, Adidas, ...}
    \item BaseColor: {Black, Red, Blue, Green, ...}
    \item Fabric: {Cotton, Polyester, Blended, ...}
    \item Priceband: {0-500, 500-1000, 1000-1500, ...., 3000+}
    \item Neck: {Round Neck, Polo Collar, V-neck, ....}
    \item Pattern: {Printed, Solid, Striped, Colorblocked, ....}
\end{enumerate}
In this approach, alongwith the product-product pairs we also generate product-SI pairs and SI-SI pairs to be input to the Word2Vec model.
For each (centre-product, context-product) pair, we generate the following tuples:
\begin{enumerate}
    \item \(({P_{centre}, P_{context}})\) 
    \item \(({P_{centre}, P_{SI_{centre}}})\), for each SI of the centre product
    \item \(({P_{centre}, P_{SI_{context}}})\), for each SI of the context product
    \item \(({P_{SI_{centre}}, P_{SI_{context}}})\), for each (SI,SI) pair from centre and context products
\end{enumerate}
By doing so we have increased vocabulary size from total number of products to total number products plus the total number of SI key-value pairs. Thus we also learn vectors for each of those key-value pair from SI.

\subsubsection{DeepWalk-Prod2Vec and DeepWalk-ProdSI2Vec}
DeepWalk was proposed as an SGNS based method for learning embeddings of nodes in a graph\cite{perozzi2014deepwalk}. They generate sequences of nodes from a graph as input sentences to Word2Vec like SGNS based architecture.

As enumerated in \cite{arora2017deciphering}, we create a weighted item-item graph. Let \(\textbf{E}_{up} = \{E_c, E_b, E_p\}\) be the set of all possible events where \(E_c = \) click event, \(E_b = \) add to bag event and \(E_p = \) purchase event. We just consider the event of highest priority for a given user and product combination with priorities defined as:
\begin{equation}
E_c <  E_b <  E_p
\end{equation}
We also define the importance score\footnote{The importance scores are computed using 1 month's data of the platform.} of an event \(e\) as 
\begin{equation} \label{eq:imp}
     I_e= 
\begin{cases}
    1,& \text{if } e = E_c\\
    \\\frac{\sum E_c} {\sum E_b},& \text{if } e = E_b\\
    \\\frac{\sum E_c} {\sum E_p},& \text{if } e = E_p
\end{cases}
\end{equation}
For \(e_{ij}\) as interaction between user \(u_i\) and product \(p_i\), we create a matrix \(W\) of dimensions \(|U| \times |P|\) where
\begin{equation} \label{eq:imp}
     w_{ij}= 
\begin{cases}
    I_e,&\text{if } e_{ij} \in \textbf{E}_{up}\\
    0,&\textit{otherwise}
\end{cases}
\end{equation}
On this matrix we apply non-negative matrix factorisation, to learn latent embeddings of user \(u\) as \(\gamma_u\) and product \(p\) as \(\gamma_p\)
Next we create a weighted item-item graph \(G = (V = P, E)\), where weight of edge between \(p_i\) and \(p_j\) is similarity between them computed as, \(\mathbf{\boldsymbol{\gamma}_{p_i} \cdot \boldsymbol{\gamma}_{p_j}} \)

Next, we use DeepWalk on this graph, to generate sequences of nodes, i.e. sequences of products. Hereafter, we refer them as \textit{simulated sessions} 

We use these simulated sessions to learn products embeddings using Prod2Vec and ProdSI2Vec as described above.

\subsection{Unified Embeddings}
We learn unified embeddings for products using embeddings learned from ProdSI2Vec, DeepWalk-ProdSI2Vec and, Images. Two sets of Unified Embeddings are generated:
\begin{enumerate}
    \item Unifying Embeddings from ProdSI2Vec and Images
    \item Unifying Embeddings from DeepWalk-ProdSI2Vec and Images
\end{enumerate}
We propose a simple weighted average to unify these embeddings: 
\begin{equation}
    \gamma_p = w_I \cdot \gamma_{p_I} + w_{PSV} \cdot \gamma_{p_{PSV}}
\end{equation}
where \(\gamma_{p_I}\) are image embeddings and  \(w_I\) is the weight associated with them, \(\gamma_{p_{PSV}}\) are Word2Vec based embeddings (ProdSI2Vec or DeepWalk-ProdSI2Vec) and \(w_{PSV}\) is the weight associated with them.
The weights are learned using grid search on the cross-validation dataset of the downstream task we use the embeddings for. For example, weights learned for Clicked-Purchased Product Similarity task (Section \ref{Res:Clicked-Purchased Product Similarity}) are 0.1 for image and 0.9 for Word2Vec based embeddings. 

\begin{figure*}[hbt!]
    \centering
    \includegraphics[width=16cm,height=8cm]{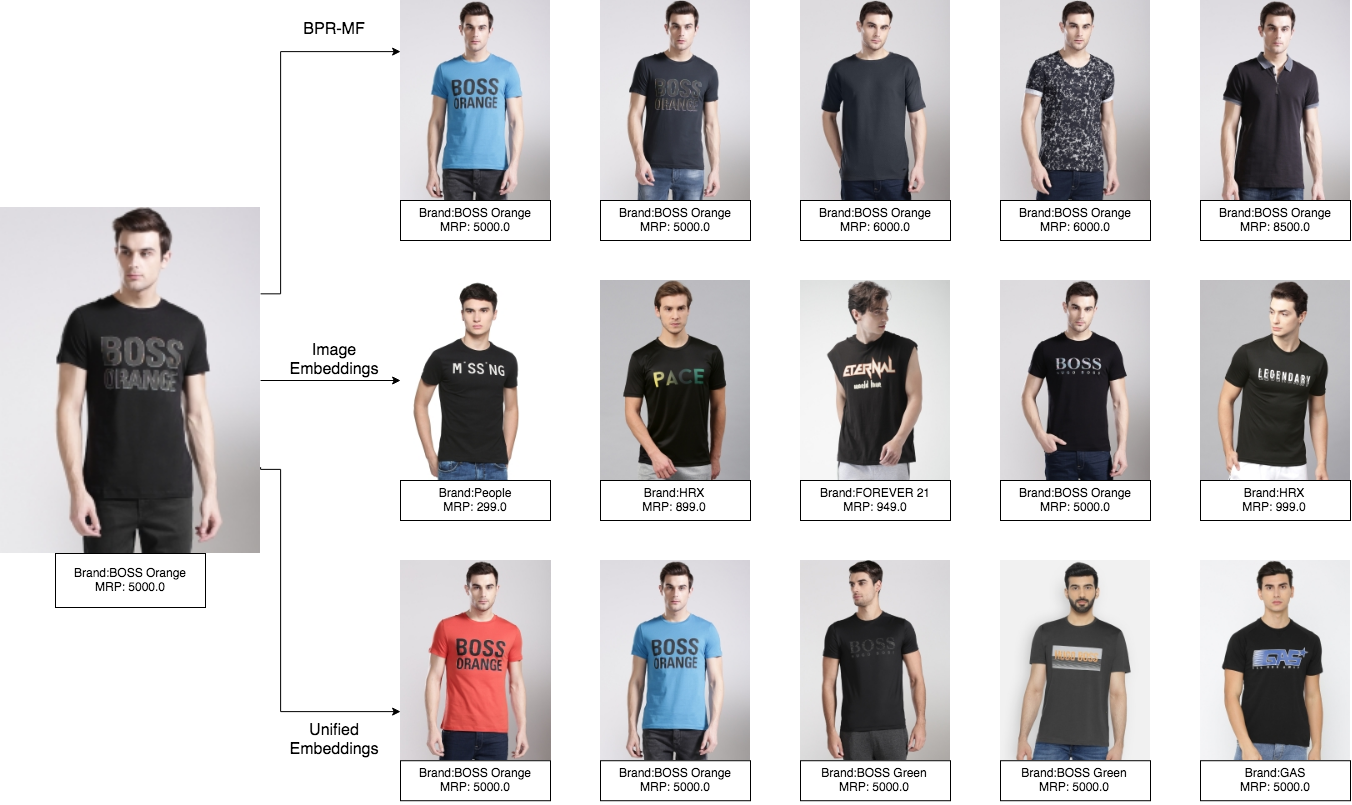}
    \caption{Most Similar Products to a given Query using different Embeddings}
    \label{fig:Qualitative_Check}
\end{figure*}

We also experimented with other unification techniques outlined in  \cite{kiela2018efficient} like additive, hadamard, max-pooling and gated aggregations for combining Word2Vec based embeddings and image embeddings. The unification approaches listed till now are generic aggregation techniques which can be reused in varied downstream tasks, however, we can also generate unified embeddings fine-tuned according to a task. For an instance, a regression model can be trained for the Cart Return Prediction task (Section \ref{Res:Cart-Return Prediction}). Input to such system would be Word2Vec based embeddings and image embeddings and output would be cart return rate. Post training, the output of the last fully connected layer could be extracted and used as unified embeddings for this task. In our experiments, we observed that embeddings generated via this technique performs marginally better for the trained task, however, performs sub-optimally for other tasks as compared to embeddings learned through generic approaches.

\subsection{Implementation}
We trained all Word2Vec based models using gensim implementation. We use default positive alpha value (negative sampling hyperparameter) of 0.75 so that popular products are sampled more often than unpopular products. Popular products have more interactions resulting in better vector representation from clickstream data. Furthermore, for products with sparse interactions side information and image embeddings provide more information. We set the window size to one because our training set consists of tuples of length 2 as explained in \ref{Word2Vec:Prod2Vec}. The vector size is set to 100 dimensions.

\section{Results} \label{results}

We evaluate the performance of all the nine embeddings on three different tasks, which chosen to be varied enough so as to be able to check the generalizability of embeddings. The generalizability of embeddings implies that they be able to capture all the signals which effect tastes of a user. Table \ref{table:Table 2} shows nine types of product embeddings which are compared.

\begin{table}[h!]
\centering
\begin{tabular}{|p{2cm}|p{3cm}|}
\hline
\textbf{Notation} & \textbf{Embedding Name}\\ \hline
BPR-MF & Bayesian Personalised Ranking Matrix Factorisation \\ \hline
DAE & Denoising AutoEncoder \\ \hline
IE & Image Embeddings \\ \hline
P2V & Prod2Vec \\ \hline
PSI2V & Prod-SI2Vec \\ \hline
DWP2V & DeepWalk-Prod2Vec \\ \hline
DWPSI2V & DeepWalk-ProdSI2Vec \\ \hline
UPSII2V & Unified-ProdSI-Images2Vec \\ \hline
UDWPSII2V & Unified-DeepWalk-ProdSI-Images2Vec \\ \hline

\end{tabular}
\caption{Product Embeddings}
\label{table:Table 2}
\end{table}

\subsection{Dataset}

We conducted our experiments on Men t-shirts category with 220k products. The users who have purchased less than 3 t-shirts on the platform are not used for generating lists of products as described in Section \ref{Word2Vec:Prod2Vec}. We randomly sampled 5M such users.

In our dataset, on an average, the number of \(E_c\), i.e. click events per user is 157, the number of \(E_b\), i.e. add to bag events per user is 19 and number of \(E_p\) i.e. purchase events per user is 6. Similarly, on an average, in a given session the number of click events is 9.3, the number of add to bag events is 2.2 and the number of purchase events is 0.53. Also, we note that the distribution of clicks of products is very long tailed, as is generally the case in an e-commerce setting. Only 20 percent of products contribute to 80 percent of clicks.

We consider only \(E_b = \) add to bag event and \(E_p = \) purchase event for P2V, DWP2V and USIIP2V. For DWP2V, DWSIP2V and UDWSIIP2V all events  \(E_c\), \(E_b \)  and \(E_p\) are considered.

\begin{figure*}
    \centering
    \includegraphics[width=12cm,height=4cm]{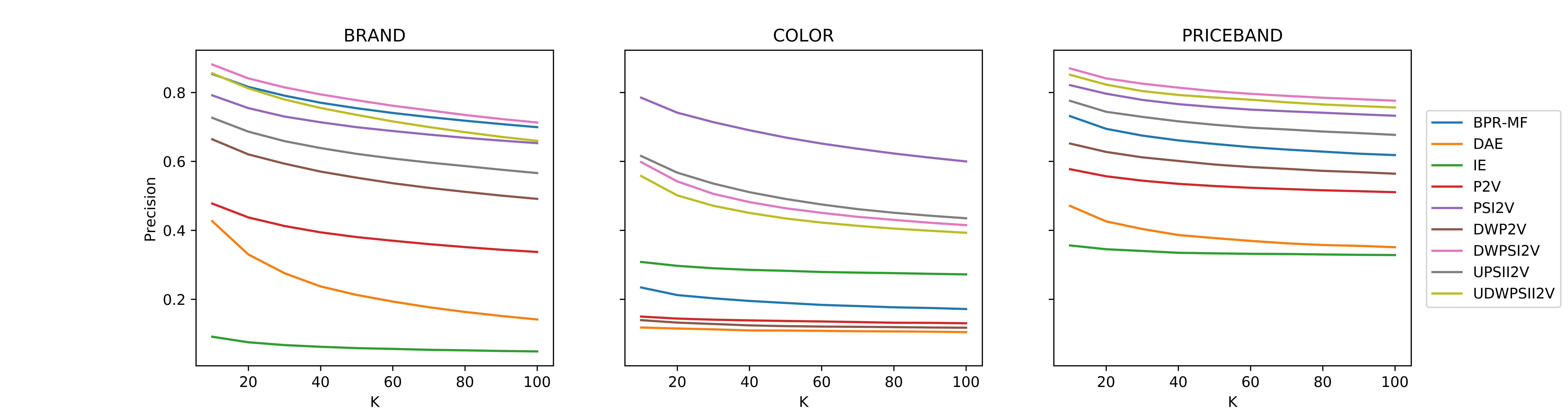}
    \caption{Precision at different values of k for different attributes}
    \label{fig:Precision @k}
\end{figure*}

\subsection{Visualisation}
In this section, we attempt to visualise our embeddings qualitatively as shown in Figure \ref{fig:Qualitative_Check}. We take a premium BOSS t-shirt as the query product. We observe that BPR-MF embeddings are biased towards the brand. However, attributes like neck and pattern are not captured well; as evident from a printed t-shirt and a polo neck t-shirt in similar products. The image-based embeddings are purely based on visual aesthetics. The similar products span multiple brands, price ranges etc. Our unified embeddings are able to capture latent product style (basis co-browsing), show similarity in terms of visual aesthetics and are also coherent in terms of attributes like brands, price, neck, pattern etc. These results were further evaluated by fashion experts in a blind study, which confirmed our hypothesis that unified embeddings are most coherent in terms of both visual aesthetics and physical attributes. 

\begin{figure}[h!]
    \centering
    \includegraphics[width=8cm,height=8cm]{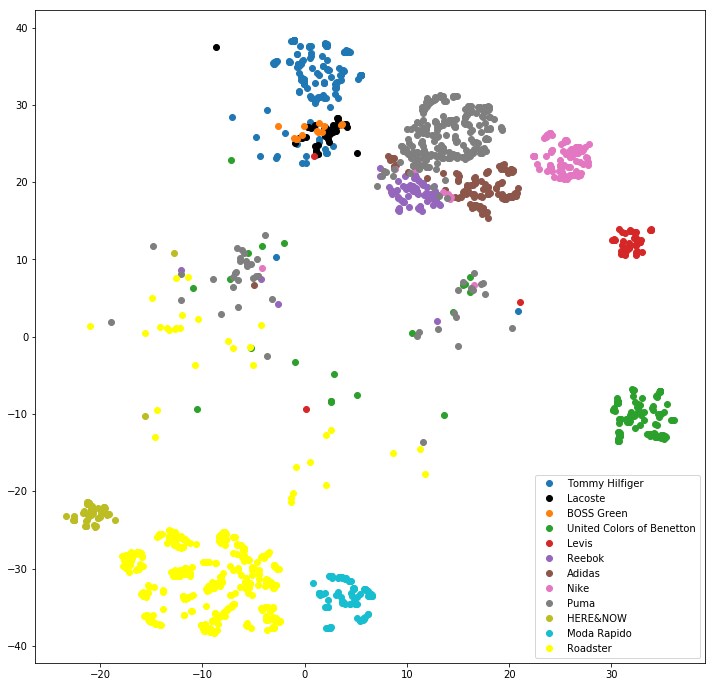}
    \caption{T-SNE plot showing Brand Clusters}
    \label{fig:brand_vis}
\end{figure}

As another experiment, we performed t-SNE on the embeddings learnt from DWPSI2V. The Figure \ref{fig:brand_vis} shows the projection of different brands in the same space.

Clearly, 4 different clusters are evident in the plot - One cluster for sports brands like Nike, Adidas, Puma and Reebok. One for casual brands like Levis and United Colors of Benetton. Another cluster includes premium brands like BOSS Green, Lacoste and Tommy Hilfiger. Finally, a cluster included a few  brands (of slightly mass-premium price range) like Roadster, Here\&Now and Moda Rapido. This clearly shows that embeddings are able to capture brand semantics fairly well so as to be able to capture user perception of brands.

\subsection{Embeddings to Attributes}
This task attempts to evaluate learnt embeddings on how well they can capture the products' textual attributes like the brand. We randomly sampled 1000 products. For each of these query products, we find top $k$ most similar products using each of our embeddings. Out of $k$ most similar products, we compute the number of products with the same brand, colour and priceband as the query product. This task tests the ability of embedding to capture these attributes.
The precision $@k$ for different values of $k$ is shown in Figure \ref{fig:Precision @k}. 

We see that image embeddings are not able to capture brand, priceband and colour information very well \footnote{Our siamese network was trained with the same coloured products as negative examples to remove high colour bias}.  As side information contains brand and colour key-value pairs, we observe that adding SI to P2V and DWP2V improves the precision substantially. Finally, as image embeddings are not able to capture brand, priceband and color information, we see that unifying them with PSI2V and DWPSI2V reduces the precision and acts more like a noise. In this task, we also observe that DeepWalk based embeddings perform better than their counterpart because of the fact that they are created using simulated cleaner sessions and hence capture brand/price/colour well. 

\subsection{Clicked-Purchased Product Similarity}
\label{Res:Clicked-Purchased Product Similarity}

A typical user's session customer involves product click events, add to bag events and purchase events. While brand, colour and price band are attributes that are easy to capture, it becomes critical to also learn the product's latent style, look and aesthetics to enable better recommendations. This task evaluates different embeddings on one such recommendation related task.

\begin{figure}[h]
    \centering
    \includegraphics[width=8cm,height=2cm]{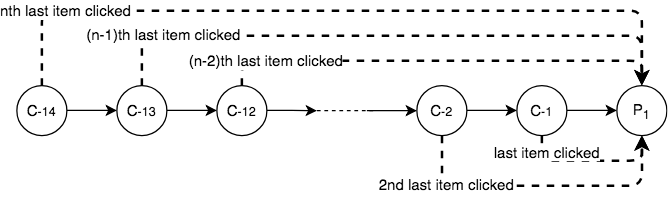}
    \caption{An example of Session considered for Evaluation of Clicked-Purchased Product Similarity Task}
    \label{fig:session_example}
\end{figure}

\begin{figure}[h!]
    \centering
    \includegraphics[width=8cm,height=5cm]{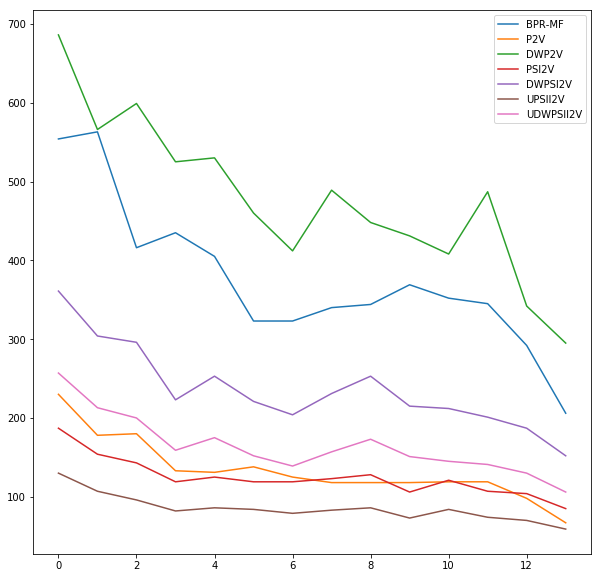}
    \caption{Median Ranks Vs. last ith product clicked}
    \label{fig:Clicked-Purchased Product Similarity}
\end{figure}

For each user who has purchased at least 3 products, we retrieve sessions as shown in Figure \ref{fig:session_example} (with a purchase) as an aggregation of last 14 clicked products and finally the purchased product. We prune the incoherent sessions by removing sessions where the median of cosine similarities between clicked and finally purchased product is less than 0.6. For each of the 14 products clicked prior to the eventual purchase, we compute the similarity rank of the products with the purchased product. Then, for a given embedding we find the rank of each product clicked in the sorted list of products most similar to the product finally purchased in that session. For this filtered set of sessions, we calculate the median of ranks for each of the last $i^{th}$ product clicked and plot a graph between median rank and the products clicked in Figure \ref{fig:Clicked-Purchased Product Similarity}. 

We clearly observe that adding SI to P2V to get PSI2V decreases the median ranks across all positions. Further unifying this with Images to get UPSII2V, the curve drops down further showing that the visual similarity contributes to the similarity between from clicked to purchased products.

Finally, we note that DeepWalk based methods don't fare very well in this task. This can primarily be attributed to the fact that DeepWalk uses much cleaner simulated sessions basis the item-item graph. However, in reality, the sessions are much noisier. Also, embeddings from pure content-based methods like Images and DAE had very high ranks, so they are not shown in the figure.

\subsection{Sparsity based experiments}

In general, most e-commerce platforms face the challenge of the sparsity of interactions with products, i.e. a large number of the products have few or no user interactions which can be leveraged to learn embeddings. This sparsity is further amplified in categories like fashion and accessories due to high invert ratio and high expiry ratio. This behaviour is also reflected in our data set.

For approaches which are based on signals from users only to learn embeddings, the vectors for sparse products are not very well in capturing item-item similarity. As we are using side information and images for creating final embeddings, for products with very fewer signals, side information and images contribute to embeddings. For this task, we do a Next Event Prediction task and report average Hit Ratio $@k$. First, we make a set of sparse products that is 45 percent products which contribute to only 5 percent of total clicks. Next, we retrieve all user sessions in which sparse products were clicked. For prediction, we use the sparse products in the session as the query product and predict the k closest products to the query product using vectors learned from different approaches. For all products clicked in a session, If the product clicked is in $k$ products then HR$@k$ for that product clicked is one otherwise zero. We report average HR$@k$ for all products in a session and for all sessions in our evaluation sets. 28k sparse products (which contribute to less than 5 percent of clicks) and 43k sessions were considered for this evaluation task.

\begin{figure}[h]
    \centering
    \includegraphics[width=8cm,height=5cm]{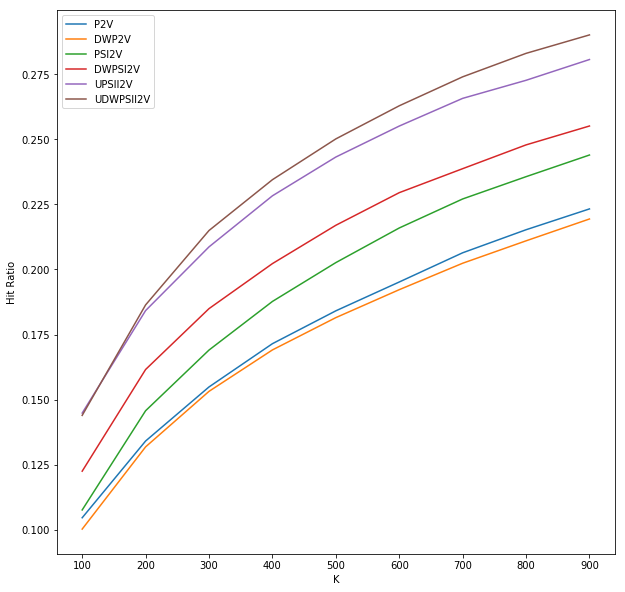}
    \caption{Hit Ratio at different K values}
    \label{fig:Sparsity}
\end{figure}

From Figure \ref{fig:Sparsity} we observe that adding side information to P2V and DWP2V improves the HR for all k values and also finally unifying the embeddings with image embeddings improves the Hit ratio further. Also here DeepWalk based variants perform better than others. We also did a very similar evaluation to test the embeddings in cold start setting, i.e. considering the products which were catalogued very recently, and found out that unification by including side information and image embeddings improves the HR$@k$.

\subsection{Cart Return Prediction}
\label{Res:Cart-Return Prediction}
Cart return prediction is unrelated downstream tasks with which evaluated our embeddings. In this task, we aim to predict users' propensity for returning product(s) from a cart at the time of purchase. Returns ensue bad user experience apart from extra operational costs incurred on the platform. As per our analysis, a product which is added to the cart is purchased 90\% of the time. Another analysis reveals that about 30\% of orders placed get returned later.  Hence, through cart return prediction, we aim to identify the cart products which have a high probability of being returned. Post identification of such products, various corrective actions like personalized shipping charges, size recommendation pop-up, Try and Buy recommendation pop-up can be undertaken to target returns. A regression-based model is used to predict whether a product in the cart will be returned or not and incorporates: (1) User attributes like returned quantity, returned revenue, online orders, purchase affinity, (2) Product attributes like product return score, price segment, (3) Order Level Features, (4) Product embedding vectors and (5) User embedding vectors. We show Test Set Precision, Recall and F1 Scores of the regression-based model in Table \ref{table:Table_5}.

\begin{table}[h!]
\centering
\begin{tabular}{|c|c|c|c|}
\hline
\textbf{Embeddings} & \textbf{Precision} & \textbf{Recall} & \textbf{F1-Score}\\ \hline
BPR-MF & 0.8 & 0.93 &	0.86 \\ \hline
DAE & 	0.8 & 0.89 & 0.84 \\ \hline
IE &	0.8 & 0.93 & 0.86 \\ \hline
P2V &	0.79 & 0.89 & 0.84 \\ \hline
PSI2V &	0.8 & 0.98 & 0.88 \\ \hline
\textbf{DWP2V} &	\textbf{0.81} & \textbf{0.98} &	\textbf{0.89} \\ \hline
\textbf{DWPSI2V} &	\textbf{0.83} & \textbf{0.96} &	\textbf{0.89} \\ \hline
UPSII2V & 0.8 &	0.98 & 0.88 \\ \hline
\textbf{UDWPSII2V} &	\textbf{0.82} &  \textbf{0.92} &  \textbf{0.87} \\ \hline

\end{tabular}
\caption{Precision and F1 Scores for Cart Return Prediction }
\label{table:Table_5}
\end{table}



We observe that DeepWalk based approaches and unified embeddings perform better than BPR-MF, DAE, IE and P2V. This implies that embeddings learned only through content or clickstream data are not generalizable enough to be used for return prediction. Hence, when we learn embeddings from side information along with clickstream data, we are able to capture the products' attributes which possibly lead to its eventual return. We also observe that unified embeddings fare very close to DeepWalk based approaches and are not better in this task which could be attributed to the fact that a product's visual aspects do not contribute much towards its return decision.

\section{Conclusion}

We propose a framework to combine multiple data sources - catalog text data, user's clickstream session data, and product images and generate a unified representation of all products in a \textit{product semantic space} . We utilized various state-of-art techniques like denoising auto-encoders for text, Bayesian personalized ranking (BPR) for clickstream data, Siamese neural network architecture for image data and a combined ensemble over the above methods for unified embeddings. Further, we compare and analyze the performance of these embeddings across three unrelated real-world e-commerce tasks like product attribute coverage, recommendation setting and predicting returns. We show that unified product embeddings perform uniformly well across all these tasks. Further, we show the efficacy of unified embeddings through experiments on three unrelated downstream tasks common to most e-commerce platforms. We observe that unified embeddings  outperform embeddings created in isolation from different data sources. As one would expect, different embeddings perform at different levels across tasks. However, relatively unified embeddings with or without DeepWalk perform optimally. Hence we propose the use of these unified embeddings in all the downstream tasks. These embeddings can be further fine-tuned specific to a task for more improvements though our experiments show that in most tasks this is marginal. Our recommendation is to use unified embeddings without DeepWalk for all personalization related tasks which involve deciphering a user's intent basis clickstream data while using DeepWalk based unified embeddings for other business tasks since they are built on much cleaner sessions capturing higher order interaction of the products.


\balancecolumns
\clearpage

%
\bibliographystyle{acm}
\bibliography{sigproc}  
%
%
\end{document}